\documentclass[runningheads]{llncs}
\usepackage{times}
\usepackage{latexsym}
\usepackage{amsmath}
\usepackage{booktabs}
\usepackage{multirow}
\usepackage{tikz}
\usepackage{inconsolata}
\usepackage[hidelinks]{hyperref}

\hypersetup{pdftitle={Question Answering over Knowledge Graphs with Neural Machine Translation and Entity Linking},pdfauthor={Daniel Diomedi, Aidan Hogan},pdfkeywords={question answering, knowledge graphs, neural machine translation, entity linking, wikidata}}

\usetikzlibrary{shapes,arrows,positioning,fit,backgrounds,matrix,chains,scopes,calc}

\newcommand{\txt}[1]{``\textit{#1}''}
\newcommand{\rdf}[1]{\texttt{#1}}

\usepackage{listings}

\colorlet{punct}{red!60!black}
\definecolor{background}{HTML}{EEEEEE}
\definecolor{delim}{RGB}{120,20,40}
\definecolor{keyw}{RGB}{0,0,192}
\colorlet{numb}{magenta!60!black}

\lstdefinelanguage{sparql}{
	sensitive=false,
	extendedchars=true,
	literate={á}{{\'a}}1 {é}{{\'e}}1 {í}{{\'{\i}}}1 {ó}{{\'o}}1 {ú}{{\'u}}1
	{Á}{{\'A}}1 {É}{{\'E}}1 {Í}{{\'I}}1 {Ó}{{\'O}}1 {Ú}{{\'U}}1
	{ü}{{\"u}}1 {Ü}{{\"U}}1 {ñ}{{\~n}}1 {Ñ}{{\~N}}1 {¿}{{?``}}1 {¡}{{!``}}1
	{<}{{{\color{delim}<}}}{1}
	{>}{{{\color{delim}>}}}{1}
	{?}{{{\color{delim}?}}}{1}
	{*}{{{\color{delim}*}}}{1}
	{+}{{{\color{delim}+}}}{1}
	{/}{{{\color{delim}/}}}{1}
	{,}{{{\color{punct}{,}}}}{1}
	{;}{{{\color{punct}{;}}}}{1}
	{.}{{{\color{punct}{.}}}}{1}
	{:}{{{\color{punct}{:}}}}{1}
	{\{}{{{\color{delim}{\{}}}}{1} {\}}{{{\color{delim}{\}}}}}{1},
	morekeywords={ask,select,from,where,order,by,distinct,limit,offset,optional,union,filter,prefix,bound,desc,regex,str,group,not,exists,minus,service,count,certain,maybe}
}

\lstdefinestyle{sparqld}{
	basicstyle=\scriptsize\ttfamily,
	identifierstyle=\color{black},
	keywordstyle=\color{keyw}\bfseries,
	ndkeywordstyle=\color{greenCode}\bfseries,
	stringstyle=\color{ocherCode}\ttfamily,
	commentstyle=\color{darkgray}\ttfamily,
	language={sparql},
	tabsize=2,
	showtabs=false,
	showspaces=false,
	showstringspaces=false,
	extendedchars=true,
	escapechar=`,
	breaklines=true,
	basewidth=0.5em,
	moredelim=[is][\color{magenta}]{~}{~},
	moredelim=**[is][\color{gray}]{£}{£},
	moredelim=**[is][\color{blue!50!black}]{$}{$},
	moredelim=[is][\color{orange!80!black}]{!}{!},
	moredelim=**[is][\color{green!50!black}]{¬}{¬},
	morecomment=[l][\color{darkgray}]{\#},
	xleftmargin=2ex,
	xrightmargin=1ex,
	aboveskip=1.5ex,
	belowskip=1.5ex
}

\lstnewenvironment{sparql}{\noindent\minipage{\linewidth}\lstset{style=sparqld}}{\endminipage}

\makeatletter
\newcommand{\sqbox}{%
	\collectbox{%
		\@tempdima=\dimexpr\width-\totalheight\relax
		\ifdim\@tempdima<\z@
		\fbox{\hbox{\hspace{-.5\@tempdima}\BOXCONTENT\hspace{-.5\@tempdima}}}%
		\else
		\ht\collectedbox=\dimexpr\ht\collectedbox+.5\@tempdima\relax
		\dp\collectedbox=\dimexpr\dp\collectedbox+.5\@tempdima\relax
		\fbox{\BOXCONTENT}%
		\fi
	}%
}
\makeatother
\usepackage{graphicx}
\begin{document}
\title{Question Answering over Knowledge Graphs with Neural Machine Translation and Entity Linking}

\titlerunning{Question Answering over Knowledge Graphs with NMT and EL}

% If the paper title is too long for the running head, you can set
% an abbreviated paper title here
%

\author{Daniel Diomedi \and
Aidan Hogan}
\authorrunning{D. Diomedi and A. Hogan}
% First names are abbreviated in the running head.
% If there are more than two authors, 'et al.' is used.
%
\institute{DCC, Universidad de Chile; IMFD\\
\email{daniel.diomedi@ug.uchile.cl}, \email{ahogan@dcc.uchile.cl}
}

\maketitle              % typeset the header of the contribution
\begin{abstract}
The goal of Question Answering over Knowledge Graphs (KGQA) is to find answers for natural language questions over a knowledge graph. Recent KGQA approaches adopt a neural machine translation (NMT) approach, where the natural language question is translated into a structured query language. However, NMT suffers from the out-of-vocabulary problem, where terms in a question may not have been seen during training, impeding their translation. This issue is particularly problematic for the millions of entities that large knowledge graphs describe. We rather propose a KGQA approach that delegates the processing of entities to entity linking (EL) systems. NMT is then used to create a query template with placeholders that are filled by entities identified in an EL phase. Slot filling is used to decide which entity fills which placeholder. Experiments for QA over Wikidata show that our approach outperforms pure NMT: while there remains a strong dependence on having seen similar query templates during training, errors relating to entities are greatly reduced.
%\keywords{First keyword  \and Second keyword \and Another keyword.}
\end{abstract}

\section{Introduction}

Knowledge graphs (KGs) adopt a graph-based abstraction of knowledge, where nodes represent entities and edges represent relations. This graph abstraction is well-suited for integrating knowledge from diverse and potentially incomplete sources at large scale. Knowledge graphs can then serve as a common substrate of knowledge for a particular organisation or community~\cite{NoyGJNPT19}. Companies using KGs include AirBnB, Amazon, eBay, Facebook, LinkedIn, Microsoft, Uber, and many more besides~\cite{kgs}. Open knowledge graphs -- including BabelNet~\cite{NavigliP12}, DBpedia~\cite{LehmannIJJKMHMK15}, Wikidata~\cite{VrandecicK14}, YAGO~\cite{HoffartSBW13} -- are made available to the public. Applications for knowledge graphs include chatbots, information extraction, personal agents, recommendations, semantic search, etc.~\cite{kgs}.

In order to query knowledge graphs, a number of structured query languages can be used, including Cypher~\cite{FrancisGGLLMPRS18} %Gremlin~\cite{Rodriguez15} 
and G-CORE~\cite{AnglesABBFGLPPS18} for querying property graphs, and SPARQL~\cite{sparql11} for querying RDF graphs. Open knowledge graphs -- most prominently DBpedia and Wikidata -- provide public query services that can receive upwards of millions of SPARQL queries per day over the Web~\cite{MalyshevKGGB18}. In order to query such services, clients must be able to formulate their queries in the SPARQL language, but most Web users are not familiar with this query language.

There thus exists a \textit{usability gap} between the information needs that non-expert users wish to express over knowledge graphs and what they can express using structured query languages. %While many works have address this gap with faceted browsers, query builders, visual tools, etc.~\cite{BikakisS16}, such systems lack expressivity in terms of what the user can request, involve multiple steps of exploration, and/or are complex for non-experts to use~\cite{FreitasCOO12}. 
Ideally a user could express their question in the natural language familiar to them and receive answers, bypassing the need to write a structured query.

A number of works have specifically addressed the task of Question Answering over Knowledge Graphs (KGQA) down through the years. Earlier works tended to use a mix of information extraction (IE) techniques in order to identify important elements of the question (entities, relations, etc.); information retrieval (IR) techniques in order to search for entities and relations in the knowledge graph that match the question; and graph exploration (GE) in order to navigate from the entities in the question along the indicated relations to find candidate answers~\cite{UngerFC14}. More recent approaches have begun to leverage neural networks for classifying questions, and ranking answers or corresponding query templates~\cite{ChakrabortyLMTLF19}. Good performance has been achieved for simple ``factoid'' questions~\cite{BordesUCW15} (such as \txt{Where was Marie Curie born?}), that can be answered by a single edge in the knowledge graph ($\rdf{Marie Curie} \xrightarrow{\rdf{born in}} \rdf{Poland}$). Handling complex questions that require joining multiple edges (such as \txt{What was the profession of Marie Curie's father?}), and/or involve query features such as counting, ordering, etc.\ (such as \txt{How many daughters did Marie Curie have?}) remains challenging.

More recent approaches have reported success in leveraging models for neural machine translation (NMT), whereby the natural language question of the user can be translated into a query in a structured language that can be evaluated directly over the knowledge graph in order to generate the desired answers (e.g., ~\cite{LukovnikovFLA17,SoruMMPVEN17,WangJLZF20,YinGR21}).

In this paper, we first discuss previous works addressing the KGQA and related tasks. We then analyse the performance of ``pure NMT'' approaches, where we find that KGQA is not as easy as it may seem: state-of-the-art NMT approaches mostly fail to generate correct queries that reflect complex user questions. We identify that the most problematic issue relates to the out-of-vocabulary problem, particularly for entities. Based on this observation, our claim is that an approach combining entity linking (EL) with NMT can improve performance versus a pure NMT approach. Specifically, NMT is used to generate an abstract query template with generic placeholders for entities, while entity linking (EL) is used to extract entities from the question and match them with the knowledge graph. Finally a slot filling (SF) technique is used to decide which entity should replace which placeholder in the query template. Our results show that although KGQA remains a challenging task, our approach greatly reduces the number of errors due to entities when compared with state-of-the-art pure NMT baselines. 

\section{Related work}

\paragraph{Related tasks} There are a number of closely related tasks to KGQA. Most notably, ``Question Answering over Linked Data'' (QALD)~\cite{UngerFC14} also targets question answering (specifically) over RDF graphs, and has been the subject of numerous works that predate the popularisation of knowledge graphs.
The Text-to-SQL~\cite{RadevKZZFRS18} task is further related to KGQA approaches that specifically try to construct a query from the natural language question, such as NMT-based approaches. In the case of Text-to-SQL, however, queries are answered over relational databases that typically do not exhibit the type of diversity present in large knowledge graphs like DBpedia or Wikidata and rather follow fixed relational schemas. Herein we primarily focus on works relating to QALD and KGQA.

\paragraph{Non-Neural Approaches} Early works in the area of QALD avoided having to deal with the full complexity of natural language by defining control languages (e.g,  GiNSENG~\cite{BernsteinKGK05}) or by using fixed templates (e.g., TBSL~\cite{UngerBLNGC12}). More flexible approaches involve using the question to guide a graph exploration (e.g., Treo~\cite{FreitasOOCS11a}, NLQ-KBQA~\cite{JungK20}), where given a question such as \txt{What was the profession of Marie Curie's father?}, such an approach may first find candidate nodes in the knowledge graph referring to Marie Curie (e.g., \rdf{db:Marie\_Curie} in DBpedia, \rdf{wd:Q7186} in Wikidata), thereafter navigating through relations matching \txt{father} (\rdf{dbo:father}, \rdf{wdt:P22}) and \txt{profession} (\rdf{dbo:occupation}, \texttt{wdt:P106}) towards a candidate answer. Further approaches (e.g., FREyA~\cite{DamljanovicACB13}) propose human-in-the-loop interaction, techniques for answering questions over multiple sources on the Web (e.g., PowerAqua~\cite{LopezFMS12}), etc. For further details on these earlier systems, we refer to the book chapter by Unger et al.~\cite{UngerFC14}.
%\section*{Acknowledgments}
%

\paragraph{Neural Approaches} More recent approaches for KGQA have begun to leverage advances in neural machine translation (NMT), rephrasing the KGQA task into one involving the translation of natural language into SPARQL. Such approaches can be equivalently viewed as being based on neural semantic parsing (NSP), as the goal is to translate natural language questions into structured query form. While neural networks have been used to classify (simple) questions into predefined categories, or to rank candidate queries, here we focus on neural translation-based approaches, referring to Chakraborty et al.~\cite{ChakrabortyLMTLF19} for other uses of neural networks.

Ferreira Luz and Finger~\cite{abs-1803-04329} propose to compute a vector representation for questions and queries that are then fed into a LSTM encoder--decoder model with an attention layer in order to perform the translation/parsing; the out-of-vocabulary problem is explicitly left open. Soru et al.~\cite{SoruMMPVEN17} propose ``Neural Semantic Machines'', which accept as input a set of templates, composed of pairs of questions and queries with entities replaced by placeholders (e.g., \txt{What was the profession of \rdf{<A>}'s father?}, with the corresponding query), and a knowledge graph, from which valid entity replacements of placeholders are computed, generating concrete instances (e.g., \txt{What was the profession of Marie Curie's father?} with the corresponding query) that are used to train an LSTM encoder-decoder model. While this approach partially addresses the out-of-vocabulary problem, tens of millions of (highly repetitive) instances would need to be used for training to cover all entities in a knowledge graph such as Wikidata. Yin et al.~\cite{YinGR21} compare a variety of NMT models in the context of KGQA, including \mbox{RNN-,} CNN- and transformer-based models, finding that the best model was the Convolutional Sequence to Sequence (ConvS2S) model~\cite{GehringAGYD17}. However, Yin et al.~\cite{YinGR21} acknowledge that the models encounter issues with larger vocabularies.

Explicitly addressing the out-of-vocabulary issue, Lukovnikov et al.~\cite{LukovnikovFLA17} propose to merge word-level and character-level representations for both the questions and the entity labels in the knowledge graph; however, the approach is specifically designed to deal with simple questions that induce a query with one \textit{triple pattern} (e.g., \txt{Who is Marie Curie's father}), and would not work with knowledge graphs that use identifiers not based on a natural language (as is the case for Wikidata). Recently Wang et al.~\cite{WangJLZF20} propose gAnswer: a system that extends NMT with entity linking (EL) and relation linking (RL) in order to better handle the out-of-vocabulary problem, finding that entity linking in particular improves performance over NMT. However, only one EL system is explored, and there is no discussion of how entities are matched with their placeholders.

\paragraph{QALD/KGQA Datasets}

Various datasets have been proposed for QALD/KGQA, including a series of QALD challenges~\cite{UsbeckGN018}, which provide a variety of small sets (in the order of hundreds) of training and test examples over knowledge graphs such as DBpedia and Wikidata. For NMT-based approaches, larger datasets are needed to ensure more complete training of the respective neural networks. A number of larger datasets have thus been proposed, where the SimpleQuestions dataset is one of the most widely used~\cite{BordesUCW15}; it contains 108,442 question--answer pairs, but focuses on simple factoid questions. The Monument dataset~\cite{SoruMMPVEN17} is based on 38 unique query templates about monuments in DBpedia, where 300--600 question--query pairs are created for each template; however, these instances focus on the data for one class in DBpedia. The DBNQA dataset~\cite{hartmann2018generating} includes 894 thousand question--query pairs produced by taking 215 queries from a QALD training dataset, replacing the entities with placeholders, and generating other valid replacements from the DBpedia knowledge graph; there is thus a high level of repetition as the 894~thousand examples are based on 215 unique queries. Finally, LC-QuAD~2.0~\cite{DubeyBA019} consists of 30 thousand question--query pairs over DBpedia and Wikidata generated from 22 templates, where the questions are paraphrased through crowdsourcing to generate more diverse questions in terms of their expression. 

\section{The Problem with Entities}\label{sec:reality}

Our general goal is to advance NMT-based approaches for the KGQA task. We thus first set out to establish a baseline for a state-of-the-art ``pure NMT'' approach by selecting three of the best performing models for the KGQA in the comparison of Yin et al.~\cite{YinGR21}. These three models constitute the state-of-the-art for NMT in the context of natural language, and represent three diverse architectures based on Recurrent Neural Networks (RNNs), Convolutional Neural Networks (CNNs) and attention without recurrence or convolutions. More specifically, the three models are as follows:

\begin{itemize}
\item ConvS2S (Convolutional Sequence-to-Sequence)~\cite{GehringAGYD17}: A CNN-based architecture, featuring gated linear units, residual connections, and attention.
\item LSTM (Long Short Term Memory)~\cite{LuongPM15}: An RNN-based architecture that uses stacking LSTM models of four layers, with attention mechanisms.
\item Transformer~\cite{VaswaniSPUJGKP17}: a more lightweight architecture that interleaves multi-head attention mechanisms with fully-connected layers.
\end{itemize}

\noindent
We follow the configurations given by Yin et al.~\cite{YinGR21}, using the Fairseq framework\footnote{\url{https://fairseq.readthedocs.io/en/latest/}} and Google Colab. We use the hyperparameters in Table~\ref{tab:hyper}. For ConvS2S, the first 9 layers had 512 hidden units and kernel width 3; the next 4 layers had 1,024 hidden units and kernel width 3; the final 2 layers had 2,048 hidden units with kernel width 1.

\begin{table}[tb]
\caption{Hyperparameter settings used~\cite{YinGR21} \label{tab:hyper}}
\centering
\setlength{\tabcolsep}{5pt}
\begin{tabular}{lrrcllrr}
\toprule
\textbf{Model} & \textbf{Layers} & \textbf{H. Units} & \textbf{Attention} & \textbf{Optimiser} & \textbf{L. Rate} & \textbf{Dropout} & \textbf{Size} \\
\midrule
ConvS2S & 15 & 512--2,048 & yes & SGD & 0.5 & 0.2 & 500 \\
LSTM & 4 & 1,000 & yes & Adam & 0.001 & 0.3 & 500 \\
Transformer & 6 & 1,024 & yes & Adam & 0.0005 & 0.3 & 500 \\
\bottomrule
\end{tabular}
\end{table}

With respect to the datasets, we first aim to replicate the results of Yin et al.~\cite{YinGR21} over the LC-QuAD~2.0 dataset, which was the most complex, diverse and challenging KGQA dataset they tested. We use the original training/test splits of LC-QuAD~2.0. However, we encountered some quality issues with the LC-QuAD~2.0 dataset, which may relate to its use of crowdsourcing platforms to generate and diversity the question texts. Specifically, participants were asked to rephrase synthetic questions such as \txt{What is [occupation] of [father] of [Marie Curie]?} into a more natural form, such as \txt{What was the occupation of Marie Curie's father?}, or to paraphrase the question, such as \txt{What did Marie Curie's father do for a living?}. From revising the benchmark, though in many cases the process yielded meaningful question--query pairs, not all cases were like this. Generating a large, diverse and high-quality benchmark dataset of complex questions for KGQA is far from trivial. In particular, we identified issues including: (1) the question text being null, empty or marked not applicable (\txt{NA}); (2) questions containing syntactic features from the synthetic question, or copying the synthetic question in a verbatim manner; (3) the question specifying the answer rather than the question; (4) the length of the question being too long or too short to be a reasonable rephrasing of the synthetic question; (5) the query containing invalid tokens. Given the number of instances in the dataset, we applied automatic heuristics to detect such cases, where we discarded 2,502 (8.2\%) of the 30,226 instances due to such quality issues. We call the resulting cleaned dataset LC-QuAD~2.0*.

Given the quality issues with LC-QuAD~2.0 dataset, and the fact that its training and test datasets feature question--query pairs based on common templates, we decided to create a novel, independent test dataset with 100 question--query pairs answerable over Wikidata. This dataset does not follow standard templates, and contains queries with a range of complexities and algebraic features, ranging from simple queries such as \txt{Who is the president of Poland}, to complex questions such as \txt{Which country that does not border Germany has German as an official language?}.\footnote{While QALD datasets may have been used, LC-QuAD~2.0 may use templates inspired by QALD, which would thus constitute an indirect leak between training and testing datasets.} We call this novel dataset WikidataQA; it contains 132 entities, 86 properties, and 208 triple/path patterns.

In Table~\ref{tab:baseline}, we present the results for the three models trained on the LC-QuAD~2.0* dataset, applied to the LC-QuAD 2.0* test set and the full WikidataQA set. These results consider a number of different metrics. First we present the BLEU score, which is a modified version of precision used to compare a given translation against a set of reference translations. This measure was reported by Yin et al.~\cite{YinGR21}. However, a translated query may have an excellent BLEU score but may be invalid or distinct from the reference query. Thus we also present accuracy, which indicates the percentage of questions for which the exact reference query was generated. This measure is itself flawed, since two queries that are syntactically different may be semantically equivalent, meaning that they return the same results. Unfortunately, determining the equivalence of SPARQL queries (or queries for any language that encapsulate first-order logic, including SQL) is undecidable~\cite{SalasH18}. A complementary measure is thus to compare the answers generated by the reference query and the computed query in terms of precision and recall; we then finally present the macro precision, recall and $F_1$-score (averaged over each question, which is given equal weight). Again however, this measure is somewhat flawed in that -- in particular for \texttt{ASK} queries that return a boolean result -- a query that is completely unrelated may return a correct result. While no measure perfectly captures KGQA performance, each provides a different insight.

\begin{table}[tb]
\caption{KGQA performance of baseline models in terms of BLEU ($B$), Accuracy ($A$), Precision ($P$), Recall ($R$), and F$_1$ score ($F_1$) for three models and two datasets\label{tab:baseline}}
\footnotesize
\centering
\setlength{\tabcolsep}{1ex}
\begin{tabular}{llrrrrrr}
\toprule
\textbf{Model} & \textbf{Dataset} & $B$ & $A$ & $P$ & $R$ & $F_1$ \\ 
\midrule
ConvS2S & LC-QuAD~2.0* & \textbf{51.5\%} & \textbf{3.3\%} & \textbf{16.4\%} & \textbf{16.6\%} & \textbf{16.4\%} \\
LSTM & LC-QuAD~2.0* & 45.6\% & 0.2\% & 12.9\% & 13.0\% & 12.9\% \\
Transformer & LC-QuAD~2.0* & 48.2\% & 2.1\% & 15.0\% & 15.2\% & 14.9\% \\
\midrule
ConvS2S & WikidataQA & \textbf{18.8\%} & 0\% & 5.5\% & 6.0\% & 5.3\% \\ 
LSTM & WikidataQA & 18.3\% & 0\% & \textbf{7.9\%} & 7.8\% & \textbf{7.9\%} \\
Transformer & WikidataQA & 18.3\% & 0\% & 7.2\% & \textbf{8.9\%} & 7.3\% \\
\bottomrule
\end{tabular}
\end{table}

The presented results indicate that while the BLEU score for LC-QuAD~2.0* is mediocre, accuracy is very poor, while precision and recall of answers is somewhat better than accuracy. We highlight that over the LC-QuAD~2.0 dataset, Yin et al.~\cite{YinGR21} report BLEU scores of 59.5\%, 51.1\% and 57.4\% for ConvS2S, LSTM and Transformer, respectively; although BLEU scores drop over our cleaned dataset for the corresponding models -- perhaps due to the removal of trivial cases -- the relative ordering of models remains the same. We can also compare these results against those reported for gAnswer~\cite{WangJLZF20} over the QALD-9 and Monument datasets, where Transformer and ConvS2S provided the best accuracy, and generally outperformed LSTM.

From the LC-QuAD~2.0* results, we thus conclude that while NMT models provide a reasonable translation, they rarely manage to compute the correct query corresponding to a user's question. The results for WikidataQA are consistently even worse than those for LC-QuAD~2.0*, indicating that when presented with questions not following a template seen in training, the models struggle to generalise.

Initially comparing the different models, we see that ConvS2S provides the best results for all metrics in the case of LC-QuAD 2.0*. However, while ConvS2S provides the best BLEU score in the case of WikidataQA, it is outperformed by LSTM and Transformer in terms of precision, recall and $F_1$.

Although the NMT-based gAnswer system won the recent QALD-9 challenge by a clear margin, these results highlight the difficulty of the KGQA task when considering more complex questions. Analysing the types of errors in more detail, we found that in 91.1\% of the cases, the set of entities in the output query did not correspond with that of the reference query, thus constituting the primary source of error found. This is perhaps to be expected, as the specific entities of a question--query pair may not have been seen during training, causing more frequent out-of-vocabulary errors than in the case of properties, which are fewer in number and easier to cover in the training set.

\section{Proposed Approach}

Entity linking (EL) identifies mentions of entities in a text and links them to the knowledge graph identifiers~\cite{WuHH18}. For example, given a text \txt{What did Marie Curie's father do for a living?}, an EL tool selecting DBpedia or Wikidata as a target knowledge graph would be expected to link \txt{Marie Curie} to \rdf{db:Marie\_Curie} or \rdf{wd:Q7186}, respectively. Some EL systems may further target common entities such as \txt{father}, linking them to \rdf{dbr:Father} and \rdf{wd:Q7565}, respectively~\cite{Rosales-MendezH19}. This can be seen as translating mentions of entities in the text to their knowledge graph identifiers as used in the query. 

Numerous EL techniques have been proposed down through the years~\cite{WuHH18}, where a common theme is to use the context of the text to disambiguate an entity. For example, in a question \txt{Who is the lead singer of Boston?}, while the mention \txt{Boston} has a higher prior probability of referring to the city, in this case it clearly refers to the rock band Boston, based on contextual clues such as \txt{singer}. A natural approach is then to delegate the handling of entities in the context of KGQA to external EL systems. This would avoid the need to train the NMT model with examples for every entity, and should also improve results versus the approach proposed for Neural Semantic Machines of basing disambiguation solely on a notion of prior probability~\cite{SoruMMPVEN17}. 

The idea of using EL in the context of KGQA is far from new~\cite{FreitasOOCS11a,DubeyBCL18,ChakrabortyLMTLF19,WangJLZF20}, but to the best of our knowledge only Wang et al.~\cite{WangJLZF20} have recently looked at incorporating EL with NMT. Their system, called gAnswer, won the QALD~7 challenge, and is thus clearly a promising approach. However, their approach is based on an individual EL system, and details are missing in terms of how cases where the EL system generates multiple entities, or where the query involves multiple entities, are handled.

\paragraph{ElNeuQA Overview} We propose an approach, called \textit{ElNeuQA}, that combines EL with NMT. Specifically, an EL system is used to identify entity mentions in the question and link them to the knowledge graph. We combine this with an NMT model that is trained and used to generate template queries with placeholders for entities. The results of the EL system can then be used to fill in these placeholders in the template query. In initial experiments, we noted some key design factors: (1) it is important that the EL system does not produce fewer entity links than there are placeholders in the template query, as this will lead to an invalid result; (2) the EL system may sometimes identify more entity links than placeholders, where the correct entities must be chosen; (3) for questions/queries with multiple entities, we must choose an entity for each placeholder.

We propose a custom ensemble EL approach, which incorporates multiple individual EL systems in order to boost recall (addressing (1)), and is combined with a voting method to rank individual entities (addressing (2)). Though various ensemble EL systems have been proposed in the literature, such as NERD~\cite{RizzoT12}, Dexter~\cite{CeccarelliLOPT13a}, WESTLAB~\cite{ChabchoubGZ16}, etc., we opted to build a custom ensemble system for two main reasons: (i) existing ensemble systems do not target Wikidata, where although it contains a large intersection of interlinked entities with knowledge-bases such as DBpedia, Wikidata, YAGO, etc., targetted by the existing systems, it also contains a much broader set of unique entities, where our ensemble thus includes an EL system targetting Wikidata; (ii) the nature of the scoring function changes per the aforementioned requirements where, rather than voting for entities to link a given mention in the text as in a typical EL scenario, we rather need to score entities to fill ``slots'' in the query.

We further propose to combine EL with a \textit{slot filling} (SF) technique that chooses which placeholders in the query to replace with which entities (addressing (3)). This slot filling component uses a sequence labeller trained on the question--query pairs to identify the role of entities in the question, matching them with their role in the query.

\begin{figure*}[t]
\centering
\newlength{\vgap}
\setlength{\vgap}{0.4cm}

\newlength{\hgap}
\setlength{\hgap}{1.5cm}

\newcommand{\hsp}{\vphantom{Ag}}

\tikzset{
	cmp/.style={ 
		draw,
		fill=gray!15,
		rounded corners=5pt,
		font=\sf\small\hsp
	},
	data/.style={ 
		draw,
		font=\scriptsize\hsp
	},
	arrin/.style={
		<-,
		latex-,
		thick
	}
}

\begin{tikzpicture}
\node[data,font=\small] (text) {\txt{How many PhD students of Marie Curie were nominated for the Nobel Prize in Chemistry?}};

\node[cmp,below=\vgap of text,xshift=0.3cm] (sl) {Sequence Labelling} edge[arrin] (text);

\node[cmp,left=1.1\hgap of sl] (el) {Entity Linking} edge[arrin] (text);

\node[cmp,right=0.2\hgap of sl] (nmt) {Neural Machine Translation} edge[arrin] (text);

\node[data,below=\vgap of sl] (slresult) {
\begin{tabular}{l@{~$\rightarrow$~}l}
\txt{Marie Curie} & \rdf{obj2} \\
\txt{Nobel Prize} & \rdf{obj1} \\
\txt{Nobel \ldots Chemistry}  & \rdf{obj1} \\
\end{tabular}
} edge[arrin] (sl);

\node[data,below=\vgap of el] (elresult) {
\begin{tabular}{r@{~}l@{~$\rightarrow$~}l}
1 & \txt{Marie Curie} & \rdf{wd:Q7186} \\
2 & \txt{Nobel \ldots Chemistry} & \rdf{wd:Q44585}\\
3 & \txt{Nobel Prize} & \rdf{wd:Q7191} \\
4 & \txt{PhD students} & \rdf{wd:Q12764792} 
\end{tabular}
} edge[arrin] (el);

\node[draw,,inner sep=0pt,below=\vgap of nmt] (nmtresult) {
\begin{minipage}{3.8cm}
\begin{sparql}
SELECT (COUNT(*) as ?ans)
WHERE {
  ?subj wdt:P1411 <obj1> .
  ?subj wdt:P184  <obj2> .
}
\end{sparql}
\end{minipage}
}  edge[arrin] (nmt);

\node[cmp,below=4.8\vgap of sl] (sf) {Slot Filling} edge[arrin] (elresult) edge[arrin] (slresult)  edge[arrin] (nmtresult);

\node[draw,inner sep=0pt,below=0.6\vgap of sf]{
\begin{minipage}{10cm}
\begin{sparql}
SELECT (COUNT(*) as ?ans)
WHERE {
  ?subj wdt:P1411 wd:Q44585 .  # nominated for the Nobel Prize in Chemistry
  ?subj wdt:P184  wd:Q7186  .  # doctoral supervisor is Marie Curie  
}
\end{sparql}
\end{minipage}}
 edge[arrin] (sf);

\end{tikzpicture}
\caption{Overview of ElNeuQA approach for an example input question (the comments in the SPARQL query are shown here for illustration purposes and are not part of the output) \label{fig:elneuqa}}
\end{figure*}
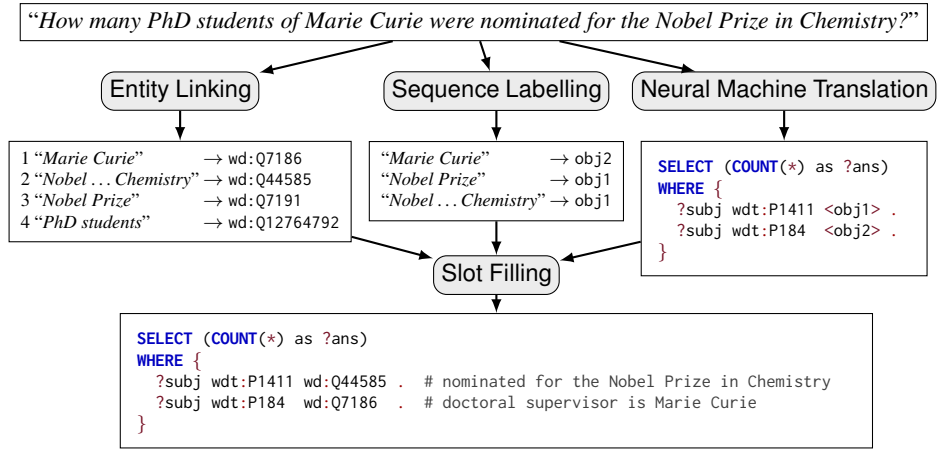

We provide an example of our proposed ElNeuQA system in Figure~\ref{fig:elneuqa}, where we see how EL, NMT and SF combine to produce the final output query. Of interest is that our EL process provides a ranked list of entities -- independent of mentions -- which is used by SF in order to prioritise the entities with the same label that fill a particular slot. In this case, NMT is used to generate a query template that leaves placeholders, such as \rdf{<obj1>} and \rdf{<obj2>} to be filled by SF. A sequence labeller is used to identify the roles of particular terms and phrases in the question, which are passed to SF. Finally, SF takes the ranked list of entities, the list of sequence labels, and the template generated by NMT, and produces the final output query replacing placeholders in the query with the highest-ranked knowledge graph entity whose mention has the corresponding sequence label. Needless to say, there exist a number of ways in which this process can give an incorrect query: EL may fail to identify all of the relevant entities; sequence labelling may produce labels that do not match the template; NMT may produce an incorrect template; etc. Later we will look at the most common forms of errors in practice.

Note that in order to train the system, the mentions of entities found in the output query must be annotated in the input question with their knowledge graph identifiers, which is the case for LC-QuAD~2.0* and our novel WikidataQA dataset.

\paragraph{Entity Linking} The EL component of ElNeuQA is instantiated with an ensemble of four EL systems: AIDA~\cite{YosefHBSW11} targetting YAGO, DBpedia Spotlight~\cite{MendesJGB11} targetting DBpedia,  OpenTapioca~\cite{Delpeuch20} targetting Wikidata, and TagME~\cite{FerraginaS10} targetting Wikipedia. Though AIDA, DBpedia Spotlight and TagME target other knowledge bases, we can use existing mappings to convert DBpedia, Wikipedia and YAGO links to Wikidata links. We primarily choose these four EL systems as they provide online APIs. Other systems could be added to the ensemble system in future. It is important for the EL component to rank the entity links that are found. Along these lines, we establish a voting system where, for a given mention, the output of each EL system lends a vote to the top-scored entity for that mention. In the case of a tie, we provide more weight to individual EL systems that provide more precise results for the training dataset; if still tied, we use the scores produced by individual systems to break the tie.

\paragraph{Sequence Labelling} The SL component is trained using the Flair\footnote{\url{https://github.com/flairNLP}} framework for NLP, and produces labels that indicate entities, strings and numbers, and their disambiguated roles (e.g., \rdf{obj2} in the case of an entity). An embedding layer is used, composed of a stacked embedding that includes GloVe embeddings and Flair contextual embeddings. The model itself follows that described by Akbik et al.~\cite{AkbikBV18}: a BiLSTM with 256 layers, and a conditional random field (CRF) layer added on top of the final layer.

\paragraph{Neural Machine Translation} The NMT component follows the baseline approach, with the sole exception that the model is trained to produce query templates with generic placeholders replacing the specific knowledge graph nodes referring to a particular entity mention (where specific nodes are rather identified by the EL component). 

\paragraph{Slot Filling} Finally, the SF component accepts the output of the previous three components and produces the final query. Specifically, it examines the placeholder labels used in the template and tries to find phrases in the question given the corresponding label by the SL component. The corresponding knowledge graph node for the phrase is selected from the EL results and used to replace each occurrence of the placeholder in the template. If multiple phrases have the same label, the ranking of the EL component is used to assign priority. In case that no phrase has a matching labelling, or a phrase with the corresponding placeholder exists but is not assigned an entity with EL, we use a ``force filling'' technique to try to replace as many placeholders as possible, knowing that unfilled placeholders will yield invalid queries. Specifically, we first apply the standard SF process as described. Next, phrases with the expected labels that overlap (rather than exactly match) with an entity mention in the EL results are used to fill placeholders. Thereafter, if a placeholder is still not filled, we assign entities that match on role, but not on order; for example, if we are searching for \rdf{obj2}, we will prioritise an unused entity labelled \rdf{obj1} over one labelled \rdf{subj1}. Finally, we will fill empty placeholders in the order they appear in the query and the EL results.

\section{Experiments}

Our hypothesis is that \textit{combining entity linking and slot filling with neural machine translation will improve the quality of results for the KGQA task}. The following experiments explore this hypothesis by extending the best three NMT models in the recent results by Yin et al.~\cite{YinGR21} and the three models used by gAnswer~\cite{WangJLZF20} -- namely ConvS2S, LSTM, and Transformer -- with the proposed EL and SF methods. We are also interested to know which parts of the ElNeuQA system cause the most errors.

\paragraph{Setting}

As per the baseline results described in Section~\ref{sec:reality}, the NMT and SL models are built with the Fairseq framework.\footnote{\url{https://fairseq.readthedocs.io/en/latest/}} We use the same hyperparameter configurations as described in Table~\ref{tab:hyper}. We will again use the cleaned LC-QuAD~2.0* for training and testing (following the original splits), and the WikidataQA dataset for testing only.

\paragraph{NMT Template Generation}

We first look at the results for generating query templates using NMT. Given that the query templates are not expected to produce answers to the question, we present only BLEU and accuracy. These results are shown in Table~\ref{tab:template}. Looking first at the results for LC-QuAD 2.0*, we can observe a moderate increase in the BLEU score versus the results for generating the full query in Table~\ref{tab:baseline} (51.5\% $\rightarrow$ 65.2\%, 45.6\% $\rightarrow$ 58.2\%, 48.2\% $\rightarrow$ 56.9\%. resp.). We also see a marked improvement in accuracy (3.3\% $\rightarrow$ 34.3\%, 0.2\% $\rightarrow$ 19.3\%, 2.1\% $\rightarrow$ 16.4\%, resp.), indicating that the models can generate query templates much more accurately than full queries mentioning specific entities, further highlighting the out-of-vocabulary issues encountered by these models when processing entities. On the other hand, the BLEU score for WikidataQA sees only a slight increase (18.8\% $\rightarrow$ 20.1\%, 18.3\% $\rightarrow$ 19.0\%, 18.3\% $\rightarrow$ 20.2\%) and accuracy remains at zero: not a single template is correctly generated by any model. This indicates that the models trained on the LC-QuAD 2.0* dataset are not capable of generalising to unseen forms of queries, as present in the WikidataQA dataset, even when entities are replaced with placeholders, highlighting the need for similarly-structured query templates to appear in the training dataset; however, when the templates are instantiated with entities, it is still possible that they return expected results in some cases, where we will later measure precision, recall and $F_1$ scores.

\begin{table}[t]
\caption{Performance of template generation in terms of BLEU ($B$) and Accuracy ($A$) \label{tab:template}}
\footnotesize
\centering
\setlength{\tabcolsep}{1em}
\begin{tabular}{llrr}
\toprule
\textbf{Model} & \textbf{Dataset} & $B$ & $A$ \\ 
\midrule
ConvS2S & LC-QuAD 2.0* & \textbf{65.2\%} & \textbf{34.3\%} \\
LSTM & LC-QuAD 2.0* & 58.2\% & 19.3\% \\
Transformer & LC-QuAD 2.0* & 56.9\% & 16.4\% \\
\midrule
ConvS2S & WikidataQA & 20.1\% & 0\% \\ 
LSTM & WikidataQA & 19.0\% & 0\% \\
Transformer & WikidataQA & \textbf{20.2\%} & 0\% \\
\bottomrule
\end{tabular}
\end{table}

\paragraph{Individual EL Systems}

Next we look at the performance of the four individual EL systems that we use for our ensemble: AIDA, DBpedia Spotlight, OpenTapioca and TagME. The results for micro and macro precision, recall and $F_1$-score are presented in Table~\ref{tab:el} over the LC-QuAD~2.0* dataset. We highlight that these results consider the entity mentions in the question text that correspond to entities in the reference query as being ``correct''; it may be the case that an EL system detects entities that would otherwise be considered valid, but are not used in the query. Systems find different trade-offs between precision and recall, where AIDA tends to be the most conservative (highest precision), while TagME tends to be the most liberal (highest recall).

\begin{table}[t]
\caption{Performance of individual EL systems for the LC-QuAD~2.0* dataset \label{tab:el}}
\footnotesize
\centering
\setlength{\tabcolsep}{0.6ex}
\begin{tabular}{lrrrrrr}
\toprule
\multirow{2}{*}{\textbf{EL System}} & \multicolumn{3}{c}{Micro} &  \multicolumn{3}{c}{Macro} \\ \cmidrule(lr){2-4}\cmidrule(lr){5-7}
 & $P$ & $R$ & $F_1$ & $P$ & $R$ & $F_1$ \\
\midrule
AIDA & \textbf{72.5\%} & 31.3\% & \textbf{43.7\%} & \textbf{38.5\%} & 30.5\% & 33.1\% \\
DBpedia Spotlight & 20.8\% & 52.7\% & 29.9\% & 23.3\% & 52.5\% & 30.8\% \\
OpenTapioca & 61.8\% & 32.0\% & 42.2\% & 34.9\% & 30.2\% & 31.4\% \\
TagME & 25.0\% & \textbf{59.5}\% & 35.2\% & 29.5\% & \textbf{59.4\%} & \textbf{37.4\%} \\
\bottomrule
\end{tabular}
\end{table}

\paragraph{Sequence Labelling}

We now look at the results for sequence labelling, where Table~\ref{tab:sl} again presents the micro and macro precision, recall and $F_1$-scores. Precision and recall tend to be balanced. However, we note that there is a significant drop in performance when considering the WikidataQA dataset; the model is trained on LC-QuAD~2.0*, where we observe that it only partially generalises to the other dataset.

\begin{table}[t]
\caption{Performance of sequence labelling for the LC-QuAD~2.0* and WikidataQA datasets \label{tab:sl}}
\footnotesize
\centering
\setlength{\tabcolsep}{0.6ex}
\begin{tabular}{lrrrrrr}
\toprule
\multirow{2}{*}{\textbf{Dataset}} & \multicolumn{3}{c}{Micro} &  \multicolumn{3}{c}{Macro} \\ \cmidrule(lr){2-4}\cmidrule(lr){5-7}
 & $P$ & $R$ & $F_1$ & $P$ & $R$ & $F_1$ \\
\midrule
LC-Quad 2.0* & 62.0\% & 66.9\% & 64.4\% & 63.8\% & 66.9\% & 64.8\% \\
WikidataQA & 22.7\% & 29.8\% & 25.8\% & 30.9\% & 31.8\% & 31.0\% \\
\bottomrule
\end{tabular}
\end{table}

\paragraph{Slot Filling} Next we test the SF component, where we specifically analyse the micro and macro precision, recall and $F_1$-scores of the pairs of the form $(l,e)$ that are generated, where $l$ is a placeholder/sequence label, and $e$ is an entity. These results are presented in Table~\ref{tab:sf}. We see that the results are the lowest thus far, which is to be expected as SF depends on the EL and SL components and thus accumulates their errors.

\begin{table}[t]
\caption{Performance of slot filling for the LC-QuAD~2.0* and WikidataQA datasets \label{tab:sf}}
\footnotesize
\centering
\setlength{\tabcolsep}{0.6ex}
\begin{tabular}{lrrrrrr}
\toprule
\multirow{2}{*}{\textbf{Dataset}} & \multicolumn{3}{c}{Micro} &  \multicolumn{3}{c}{Macro} \\ \cmidrule(lr){2-4}\cmidrule(lr){5-7}
& $P$ & $R$ & $F_1$ & $P$ & $R$ & $F_1$ \\
\midrule
LC-Quad 2.0* & 47.2\% & 45.6\% & 46.4\% & 49.6\% & 47.4\% & 47.4\% \\
WikidataQA & 18.0\% & 17.0\% & 17.5\% & 18.3\% & 21.5\% & 18.8\% \\
\bottomrule
\end{tabular}
\end{table}

\paragraph{Query Generation and Question Answering} We look at the BLEU and accuracy results for the query generation, as well as the macro precision, recall and $F_1$-score for question answering in terms of the results generated. We present the results in Table~\ref{tab:final}, where the baseline ``pure NMT'' results for ConvS2S, LSTM and Transformer are the same as were presented previously in Table~\ref{tab:baseline}. In terms of models, we see that ConvS2S clearly outperforms the other two. We see that although correct answers are generated in some cases, the accuracy for WikidataQA remains at zero (a result implied by Table~\ref{tab:template} since an incorrect template implies an incorrect final query); our conclusion is that the models trained on LC-QuAD~2.0* do not generalise to the WikidataQA dataset.\footnote{We further tried a variant of ElNeuQA that considers the first query returning some result in the top five generated, which improved accuracy for LC-QuAD~2.0* up to 20.7\% and $F_1$ up to 29.6\%; however, the accuracy for WikidataQA remained zero.} Aside from this result, ElNeuQA variants achieve better results for \textit{all} models and metrics compared to the baseline of their pure NMT counterparts. Based on the results over both datasets for three state-of-the-art NMT models, we thus claim that the combination of EL, SF and NMT -- as proposed herein -- outperforms pure NMT on the KGQA task.

\begin{table}[tb]
\caption{KGQA performance of baseline and ElNeuQA in terms of BLEU ($B$), Accuracy ($A$), Precision ($P$), Recall ($R$), and F$_1$ score ($F_1$) for LC-QuAD~2.0* and WikidataQA \label{tab:final}}
\footnotesize
\centering
\setlength{\tabcolsep}{0.7ex}
\begin{tabular}{llrrrrrr}
\toprule
\textbf{System} & \textbf{Dataset} & $B$ & $A$ & $P$ & $R$ & $F_1$ \\ 
\midrule
ConvS2S & LC-QuAD~2.0* & 51.5\% & 3.3\% & 16.4\% & 16.6\% & 16.4\% \\
ElNeuQA-ConvS2S & LC-QuAD~2.0* & \textbf{59.3\%} & \textbf{14.0\%} & \textbf{26.9\%} & \textbf{27.0\%} & \textbf{26.9\%} \\[1ex]
LSTM & LC-QuAD~2.0* & 45.6\% & 0.2\% & 12.9\% & 13.0\% & 12.9\% \\
ElNeuQA-LSTM & LC-QuAD~2.0* & 52.9\% & 7.0\% & 20.2\% & 20.3\% & 20.2\% \\[1ex]
Transformer & LC-QuAD~2.0* & 48.2\% & 2.1\% & 15.0\% & 15.2\% & 14.9\% \\
ElNeuQA-Transformer & LC-QuAD~2.0* & 51.8\% & 7.1\% & 19.9\% & 20.1\% & 19.8\% \\\midrule
ConvS2S & WikidataQA & 18.8\% & 0\% & 5.5\% & 6.0\% & 5.3\% \\ 
ElNeuQA-ConvS2S & WikidataQA & \textbf{20.5\%} & 0\% & \textbf{12.9\%} & \textbf{12.9\%} & \textbf{12.9\%} \\[1ex]
LSTM & WikidataQA & 18.3\% & 0\% & 7.9\% & 7.8\% & 7.9\% \\
ElNeuQA-LSTM & WikidataQA & 18.8\% & 0\% & 8.9\% & 8.9\% & 8.9\% \\[1ex]
Transformer & WikidataQA & 18.3\% & 0\% & 7.2\% & 8.9\% & 7.3\% \\
ElNeuQA-Transformer & WikidataQA & 19.8\% & 0\% & 11.9\% & 11.9\% & 11.9\% \\
\bottomrule
\end{tabular}
\end{table}

\paragraph{Sources of error}

Although ElNeuQA has been found to improve upon state-of-the-art (pure) NMT baselines in the case of both KGQA datasets, the metrics leave much to be desired. In order to better understand these limitations, we analyse the main sources of error found. Since the accuracy results for WikidataQA were 0\%, in Table~\ref{tab:error} we present the error rates for the LC-QuAD~2.0* dataset, where T refers to a correct template, E refers to the correct entities, and S refers to correct slot filling (we skip sequence labelling, which is implicit in slot filling). We use ! to denote an error in the corresponding step and omit both T!S and !!S as an error in E implies an error in S. Of note is that all three pure NMT baselines produce a much lower percentage of queries containing correct entities versus ElNeuQA; for example, comparing ConvS2S with ElNeuQA-ConvS2S, only 10.0\% of the queries produced by the former contain correct entities, while 58.2\% of the latter contain correct entities. This indeed suggests that ElNeuQA gains most over pure NMT approaches in terms of how it handles entities. Focusing specifically on ElNeuQA, template errors were the most common. 

\begin{table}[tb]
\caption{Sources of error for in the LC-QuAD 2.0* dataset  \label{tab:error}}
\footnotesize
\centering
\setlength{\tabcolsep}{0.7ex}
\begin{tabular}{lrrrrrr}
\toprule
\textbf{System} & TES & TE! & T!! & !ES & !E! & !!!  \\ 
\midrule
ConvS2S & 3.3\% & 1.3\% & 25.3\% & 2.8\% & 2.6\% & 64.9\% \\
ElNeuQA-ConvS2S & 14.0\% & 4.0\% & 13.5\% & 15.7\% & 24.5\% & 28.5\% \\[1ex]
LSTM & 0.2\% & 0.1\% & 20.3\% & 0.7\% & 0.6\% & 78.1\% \\
ElNeuQA-LSTM & 7.9\% & 2.5\% & 8.2\% & 21.4\% & 27.3\% & 33.7\% \\[1ex]
Transformer & 2.1\% & 1.0\% & 22.1\% & 4.0\% & 4.3\% & 66.6\% \\
ElNeuQA-Transformer & 7.1\% & 1.7\% & 6.8\% & 17.6\% & 31.6\% & 35.1\% \\
\bottomrule
\end{tabular}
\end{table}

\section{Conclusions}

Question Answering over Knowledge Graphs (KGQA) has seen significant advances in recent years, but more complex questions remain a major challenge. Recent approaches have demonstrated promising results using neural machine translation (NMT) techniques. However, NMT suffers from the out-of-vocabulary problem, particularly in the case of entities, where millions of training examples would be required to cover all entities in large knowledge graphs. We thus propose an approach that we call ElNeuQA, which combines entity linking (EL) with NMT, where slot filling (SF) is used to handle complex cases involving multiple entities. Our experiments show that combining EL with NMT outperforms a pure NMT approach for all models and datasets considered. These results -- along with the recent results of Wang et al.~\cite{WangJLZF20} -- highlight the potential for NMT and EL to complement each other, where EL alleviates the out-of-vocabulary issues associated with the millions of entities that large knowledge graphs describe.

Focusing on the limitations of our work, the results for more complex questions -- particularly questions following previously unseen templates -- remain quite poor. KGQA considering such questions is thus clearly a challenging task, but one that also has the potential to make knowledge graphs accessible to a much wider audience. Regarding future work, we identified some quality issues with the LC-QuAD~2.0 dataset, where NMT-based approaches may perform better given larger, higher-quality datasets for training; however, creating such datasets is a costly exercise. Another interesting avenue to explore is to extend NMT with relation linking (RL), as proposed by Wang et al.~\cite{WangJLZF20}; however, sometimes a query relation is not explicitly mentioned by a query, and while large knowledge graphs typically contain millions of entities, they only contain thousands of relations (or properties), meaning that it would be more feasible to cover all such relations in the training data. Currently, NMT approaches will often generate queries that yield empty results, where it would be of interest to see how data from the knowledge graph could be leveraged in order to avoid generating such queries; however, user questions may indeed sometimes reflect queries with empty results, where caution would be required to avoid ``forcing'' a query with results in such cases.

\medskip\noindent
\textit{Supplementary material, including code and queries, are available from the repository \url{https://github.com/thesemanticwebhero/ElNeuKGQA}.}

\bibliographystyle{splncs04}
\bibliography{references}

\end{document}